\documentclass[letterpaper]{article} 
\usepackage{aaai2027}  
\usepackage[hyphens]{url}  
\usepackage{graphicx} 
\usepackage{natbib}  
\usepackage{caption} 
\usepackage{algorithm}
\usepackage{algorithmic}

\usepackage{newfloat}
\usepackage{listings}
\DeclareCaptionStyle{ruled}{labelfont=normalfont,labelsep=colon,strut=off} 
\floatstyle{ruled}
\newfloat{listing}{tb}{lst}{}
\floatname{listing}{Listing}

\usepackage{booktabs}
\newcommand{\method}{\textsc{Co-VStream}}

\newcommand{\cf}{\textit{cf}. }
\newcommand{\eg}{\textit{e}.\textit{g}.}
\usepackage{bm}
\usepackage{amssymb}
\usepackage{mathtools} 
\usepackage{multirow}
\usepackage{colortbl}
\makeatletter
\def\copyright@text{}
\makeatother

\usepackage[colorlinks=true, citecolor=black, linkcolor=blue]{hyperref}

\title{\method{}: Edge-Cloud Collaboration for \\ Understanding of Long Video Streams}
\author{
    Xu Liu\textsuperscript{\rm 1},
    Guikun Chen\textsuperscript{\rm 1},
    Zihao Yan\textsuperscript{\rm 2}, 
    Kanzhi Wu\textsuperscript{\rm 2}, 
    Wenguan Wang\textsuperscript{\rm 1}\corresponding
}
\affiliations{
    \textsuperscript{\rm 1}The State Key Lab of Brain-Machine Intelligence, Zhejiang University\\
    \textsuperscript{\rm 2}vivo Mobile Communication Co., Ltd., Shenzhen, China.\\

}

\begin{document}

\maketitle

\begin{abstract}
Long, continuous video streams are an increasingly critical driver of multimedia intelligence. Existing efforts often handle long videos with a sample–encode–reason approach using large models. However, they overlook a crucial deployment fact: the stream is often produced by computationally constrained devices. {This forces an untenable compromise: cloud offloading unlocks strong reasoning but incurs prohibitive bandwidth overhead, while on-device processing remains limited by edge hardware capacity.}
Therefore, we propose \method{}, the first \textit{edge-cloud collaborative} framework for understanding long video streams. The edge node distills raw video streams into compact visual features and semantic captions for transmission to the cloud, minimizing bandwidth costs, while the cloud server integrates this data into an entity graph and global visual context, activating the heavy reasoning model only when a {user query arrives. Experiments on VideoMME-Long, LVBench, and RTV-Bench show that} \method{} reduces bandwidth usage by \textbf{87.6\%} while retaining \textbf{99.2\%} of the cloud baseline accuracy on LVBench.
\end{abstract}

\section{Introduction}
\noindent\textbf{Background.}
The proliferation of always-on devices, such as AI glasses, has elevated long, continuous video streams as a critical modality~\citep{ego4d,project,egolife,longsurvey,hourvideo}. With richer temporal context, models may solve episodic memory tasks~\citep{mem0, mem1}, such as recalling past interactions or finding lost items~\citep{where,naq,egoground,egoschema}. Consequently, effectively processing the continuous data is essential for unlocking new possibilities in personalized assistance, long-term healthcare monitoring, \textit{etc}.

\begin{figure}[t]
    \centering
    \includegraphics[width=1.0\linewidth]{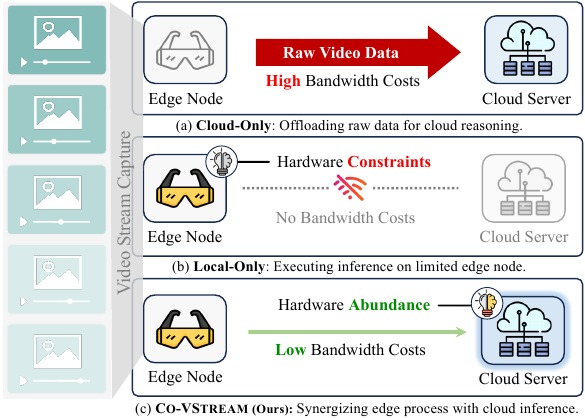}
    \caption{Comparison of deployment constraints for long video understanding on computationally limited devices.}
    \label{fig:intro}
\end{figure}

\noindent{\textbf{Motivation.}}
Existing efforts mainly handle long videos using a sample-encode-reason approach~\citep{MovieChat+,LLaMA-VID,SEAL,timechat}, which relies on heavy, centralized, large multimodal models (LMMs)~\citep{LWM,LongVILA,Video-CCAM,Video-XL}. While effective in offline benchmarks, such methods ignore a fundamental deployment constraint: the streams originate from computationally constrained edge devices (\eg, smart glasses). This disconnect forces an untenable compromise.
{\textit{Cloud-centric offloading} (Fig.~\ref{fig:intro}a) unlocks powerful reasoning but necessitates continuous raw-video transmission, causing prohibitively high bandwidth cost. Conversely, \textit{on-device processing} (Fig.~\ref{fig:intro}b) preserves bandwidth but is strictly bound by tight hardware limitations, preventing the deployment of the heavy models typically required for reasoning. Therefore, there is an urgent need for a collaborative framework that resolves this dilemma, reconciling the reasoning capacity of the cloud with the strict bandwidth constraints of edge deployment.}

\noindent\textbf{Methodology.}
We address this challenge by rethinking the spatiotemporal redundancy inherent in video streams. We posit that to resolve memory-intensive queries, compact semantic representation serves as a sufficient surrogate for raw pixels. This insight suggests an optimal division of labor: the edge acts as a semantic compressor to filter redundancy, enabling the cloud to focus on long-term memory maintenance and on-demand reasoning.
{To implement this, we propose the training-free \method{} (Fig.~\ref{fig:intro}c), the \textbf{first} edge-cloud collaborative framework designed for long video streams.} On the edge, a Dual-Condensed Perception Pipeline (\S\ref{method:edge}) processes the stream into two modalities in parallel: condensed visual features and semantic captions. {These compact payloads are uploaded to the cloud, which continuously integrates them into a global visual context and an entity graph (\S\ref{method:cloud}). }The reasoning model remains dormant until triggered by a user query. This design ensures that high-cost computation is incurred only when necessary.

\noindent{\textbf{Merits.}}
{Through this collaborative design, \method{} offers four advantages. First, it is \textit{bandwidth-efficient}. The dual-condensation pipeline drastically reduces data transmission requirements compared to raw streaming, enabling sustainable long-duration operation characterized by a low memory footprint and bandwidth costs (\S\ref{sec:exp}). Second, it is \textit{cloud-empowered}. Unlike edge-only solutions limited by hardware, it leverages cloud-side memory and reasoning to achieve performance comparable to Cloud-Only. Third, it is \textit{real-time responsive}. By decoupling memory management from reasoning, the cloud maintains the memory without blocking, enabling instantaneous query responses with latency lower than the Cloud-Only. Finally, it is \textit{training-free}. By performing all data processing in the LMM's original feature space, preserving the original semantic distribution, \method{} allows for plug-and-play integration with various LMMs.}

\noindent{\textbf{Results and Analysis.}}
{We evaluate \method{} on three streaming-oriented benchmarks, VideoMME-Long, LVBench, and RTV-Bench, comparing it against three baselines: Local-Only (\texttt{LO}), Cloud-Only (\texttt{CO}), and Edge-Cloud (\texttt{EC}). The results show that \method{} balances performance, bandwidth, and latency.} Specifically, it reduces bandwidth consumption by \textbf{87.59\%} compared to \texttt{CO} and achieves a latency of \textbf{2.99s}, improving over \texttt{CO} (4.08s), while retaining \textbf{99.2\%} of \texttt{CO} upper-bound accuracy on LVBench (\S\ref{exp:main}). 
Crucially, \method{} exhibits robustness in ultra-long video tests (\textgreater 6,000s). It surpasses \texttt{CO} by \textbf{7.92\%}. Furthermore, a 24-hour continuous stream simulation reveals a \textbf{523$\times$} reduction in memory footprint (stabilizing at 439 MB vs. 230 GB for all baselines), verifying its viability for always-on deployment on mobile agents  (\S\ref{exp:ana}). {We also include several comprehensive additional evaluations on open-ended QA, payload formatting, network stress tests, and bank-capacity ablations in the Appendix.}


\section{Methodology}
\subsection{System Overview}
\label{sec:overview}
\method{} is formalized as a dual-end collaborative system comprising an edge node and a cloud server (\cf Fig.~\ref{fig:intro}). The workflow originates at the edge, which transforms video streams into semantic payloads and user questions into query triggers. These transmissions drive video memory management and instant reasoning on the cloud.

\noindent\textbf{Edge Node.} The edge acts as the perception unit, executing two concurrent processes: \textbf{i}) Perception Stream. The edge continuously acquires the video stream and maps the raw frames into a semantic payload $\mathcal{P}_{load}$ to mitigate redundancy. $\mathcal{P}_{load}$ integrates condensed visual features $\bm{Z}_{cond}$ and keyframe captions $\mathcal{C}_{text}$. \textbf{ii}) Query Monitoring. The edge listens for user query $Q$, which serves as a wake-up signal. Upon receiving $Q$, the edge transmits a single trigger to the cloud, activating the dormant inference module in parallel without interrupting the perception stream.

\noindent\textbf{Cloud Server.}
The cloud is modeled as two decoupled, parallel processes: \textbf{i}) Memory Management. It continuously integrates the incoming $\mathcal{P}_{load}$ to dynamically update both a global visual context ($\bm{H}$) and a semantic entity graph ($\mathcal{G}$). \textbf{ii}) On-Demand Reasoning. A process remains in sleep mode by default to conserve resources, switching to active mode only upon receiving a trigger which is sent by the edge node. It then generates a response $R$ to the user query $Q$ based on \method{}'s current memory.

\begin{figure*}[t]
    \centering
    \includegraphics[width=0.99\linewidth]{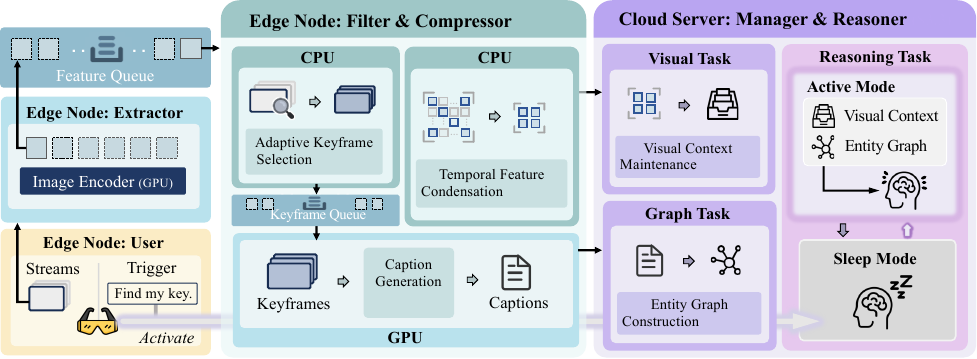}
     \put(-139.5,16){\scriptsize
    (Eq. \ref{eq:ev}-\ref{eq:cv})}
    \put(-236,169.7){\textbf{\footnotesize
    (\S\ref{method:edge})}}
    \put(-32,170){\textbf{\footnotesize
    (\S\ref{method:cloud})}}
    \put(-38,137.5){\scriptsize
    (Eq.~\ref{eq:gs})}
    \caption{Overview of \method{} (\S\ref{sec:overview}).}
    \label{fig:overview}
\end{figure*}

\subsection{{Edge Node: Dual-Stream Perception and Condensation Pipeline}}

\label{method:edge}
Given the limited resources on edge devices and the high computational cost of neural networks, serial execution of perception tasks inevitably leads to latency accumulation, rendering real-time video stream processing unfeasible. To mitigate this bottleneck, we design a cascaded pipeline that processes data across asynchronous stages.
The pipeline initiates with \textit{Visual Feature Extraction} (Eq.~\ref{eq:vis_enc}), which processes the video stream and pushes extracted features into a primary \textit{Feature Queue}. 
Subsequently, these features are retrieved to concurrently execute \textit{Temporal Feature Condensation} (Eq.~\ref{eq:condense}) and \textit{Adaptive Keyframe Selection} (Eq.~\ref{eq:keyframe}), where the latter pushes the selected keyframes into a secondary \textit{Keyframe Queue}.
Finally, this queue feeds \textit{Semantic Caption Generation} (Eq.~\ref{eq:captions}), acting as the terminal consumer to produce captions. To explain the internal mechanism of this pipeline, details are elaborated below:

\noindent{\textbf{Dual-Condensed Perception Pipeline.}}
{It transforms the video stream into compact semantic payloads through the following four operations:}

\noindent$\bullet$ \textit{Visual Feature Extraction.} 
The visual encoder $\mathcal{E}_{vis}$ extracts a feature vector $\bm{f}_t$ for each frame $v_t$ in the video stream, and buffers it into the \textit{Feature Queue}:
\begin{equation}
\bm{f}_t = \mathcal{E}_{vis}(v_t), \quad \bm{f}_t \in \mathbb{R}^{N \times d},
\label{eq:vis_enc}
\end{equation}
where $N$ denotes the number of visual patches and $d$ denotes the embedding dimension. To mitigate the spatiotemporal redundancy contained within the raw feature stream, the following condensation and filtering mechanism is required.

\noindent$\bullet$ \textit{Temporal Feature Condensation.} 
The features are continuously retrieved from the \textit{Feature Queue} to populate a condensation buffer. Only when this buffer reaches its full capacity $N_{temp}$ is the iterative merging algorithm triggered. The algorithm recursively calculates cosine similarities between adjacent feature pairs, identifies the pair with the maximum similarity, and merges them into a mean feature:
\begin{equation}
j^* = \operatorname*{argmax}_{i} \cos(\bm{f}_i, \bm{f}_{i+1}), \quad \bm{f}_{j^*} \coloneqq \frac{\bm{f}_{j^*} + \bm{f}_{j^*+1}}{2}.
\label{eq:condense}
\end{equation}
This iteration continues until the feature count is reduced to a pre-set quota $N_{target}$, yielding a set of condensed visual features $\bm{Z}_{cond}$. At this stage, the raw video stream is distilled into condensed features for cloud transmission. This process achieves a compression rate of 87.6\% (Table~\ref{tab:ablation}).

\noindent$\bullet$ \textit{Adaptive Keyframe Selection.} 
Concurrently, the features in the \textit{Feature Queue} are continuously dequeued to populate a selection buffer. Once the buffer accumulates $N_{w}$ frames, an iterative clustering algorithm is triggered to filter out the redundancy of raw features. Starting with $k = \lfloor\sqrt{N_{w}}\rfloor$ clusters, the algorithm iteratively merges the nearest cluster pairs to find the structure that maximizes the silhouette coefficient. The frames closest to the cluster centers of this optimal configuration are selected as the keyframe set $\mathcal{K}$:
\begin{equation}
    \mathcal{K} = \texttt{Select}(\{\bm{f}_i\}_{i=1}^{N_{w}}) = \{k_1, k_2, ..., k_m\},
\label{eq:keyframe}
\end{equation}
where $m \ll |N_{w}|$.
Upon selection, the keyframes are pushed into the \textit{Keyframe Queue}, which subsequently triggers the downstream semantic caption generation task.

\noindent$\bullet$ \textit{Semantic Caption Generation.} Once triggered by the arrival of the keyframes, a lightweight language model~\citep{qwen3} is employed to generate a frame caption $\mathcal{C}_{text}$ for each frame stored in the \textit{Keyframe Queue}:
\begin{equation}
    \mathcal{C}_{text} = \{c_i | c_i = \texttt{Caption}(k_i), \forall k_i \in \mathcal{K}\}.
\label{eq:captions}
\end{equation}
We enforce a ``strict simple sentences'' constraint via the prompt engineering to minimize parsing ambiguity for the cloud-side graph construction. 

{The final payload $\mathcal{P}_{load} = \{\bm{Z}_{cond}, \mathcal{C}_{text}\}$ is transmitted to the cloud. Raw video frames are converted into compact features and captions for upload. To enable the real-time execution of this pipeline on constrained edge nodes, we orchestrate a hardware scheduling:}

\noindent\textbf{Hardware-Aware Parallel Scheduling.} 
A hardware affinity partitioning strategy is implemented to execute the above pipeline on heterogeneous edge hardware by mapping each operation to a suitable processor. \textit{Visual Feature Extraction} and \textit{Semantic Caption Generation} involving dense matrix calculations and requiring high parallel throughput are assigned to the GPU. \textit{Adaptive Keyframe Selection} and \textit{Temporal Feature Condensation}, relying on dynamic clustering and iterative calculations, are assigned to the CPU. This hardware-aware mapping eliminates resource contention and maximizes the overall system throughput.

\noindent\textbf{Merits.} The cascaded design decouples the computational loads of different modules, ensuring that their execution does not mutually interfere, thereby maximizing \method{} efficiency. The hardware scheduling mitigates resource contention by directing tasks to their suitable processors. Specifically, a naive serial implementation requires 64.98ms to process each frame, whereas \method{} reduces it to 52.03ms. Collectively, these optimizations maximize \method{} efficiency, enabling real-time perception even on resource-constrained edge nodes.

\subsection{Cloud Server: Decoupled Management and Reasoning Architecture}
\label{method:cloud}
The cloud server is tasked with concurrently processing continuous semantic streams and providing timely responses to user queries. A serial architecture would allow memory updates to block reasoning, causing high latency. We thus propose a multi-process asynchronous architecture where \textit{Memory Management} continuously updates \method{}'s memory, while \textit{On-demand Reasoning} remains dormant until triggered to instantaneously generate responses.

\noindent\textbf{Memory Management.}
This process persistently listens to and decomposes the payload $\mathcal{P}_{load}$ into visual and textual components to update \method{}'s memory:

\noindent$\bullet$ \textit{Global Visual Context Maintenance.}
It receives visual component $\bm{Z}_{cond}$ to maintain the global visual context $\bm{H}_t$. To prevent memory saturation under infinite video streaming conditions, we employ the iterative merging algorithm to manage a compression bank with a capacity limit $N_{bank}$. While the global visual context provides a condensed visual history, it loses fine-grained details. To compensate for this deficiency, we parallelly process the textual component $\mathcal{C}_{text}$ to build a structured entity graph in the cloud.

\noindent$\bullet$ \textit{Real-time Entity Graph Construction.}
It incrementally transforms the unstructured captions $\mathcal{C}_{text}$ into a structured, dynamic Entity Graph $\mathcal{G}=(\mathcal{V}, \mathcal{E})$ in real-time. The construction is executed via a three-stage pipeline:

\noindent\textbf{i}) \textit{Syntactic Parsing}: We first deconstruct captions into Subject-Verb-Object (SVO) triplets. For instance, ``A man sits on the bench'' is parsed into the triplet (man, sit, bench). Subsequently, we map these components to graph elements. The extracted subjects and objects are instantiated as entity nodes ($\mathcal{V}$), while the Verbs are modeled as directed edges ($\mathcal{E}$) connecting the source and target nodes.

\noindent\textbf{ii}) \textit{Dynamic Entity Fusion}: However, raw triplets often suffer from inconsistent entity references across time. To resolve this, a \textit{Temporal Candidate Pool} $\mathcal{P}_{temp}$ containing entities detected within the last 30-second window is maintained. We employ a Natural Language Inference (NLI) classifier for bidirectional entailment verification~\citep{bientail}, where the output space is \textit{\{Entailment, Contradiction, Neutral\}}. For a new entity $E_{new}$ and a candidate entity $E_{old} \in \mathcal{P}_{temp}$, they are fused into a single node if and only if the bidirectional logical implication holds:
\begin{equation}
    \!\texttt{NLI}(E_{new}, E_{old}) \!=\!  \texttt{NLI}(E_{old}, E_{new}) \!=\! \textit{Entailment}. 
\label{eq:ev}
\end{equation}
Otherwise, instantiate $E_{new}$ as a new node, maintaining entity continuity without redundant nodes.

\noindent\textbf{iii}) \textit{Vectorized Node Mounting}: {Finally, we adopt an embedding model $\mathcal{E}_{txt}$ to encode each caption $c \in \mathcal{C}_{text}$ into a vector and attach it to the corresponding entity node for efficient natural language retrieval.}:
\begin{equation}
    \mathbf{e} = \mathcal{E}_{txt}(c), \quad \forall c \in \mathcal{C}_{text}.
\label{eq:cv}
\end{equation}
Consequently, each node in $\mathcal{G}_t$ enables rapid retrieval based on similarity in the reasoning phase.

Through the management, \method{} maintains an up-to-date memory. To ensure timely responses, the following reasoning module remains decoupled from this memory management, engaging only when triggered by the users.

\noindent\textbf{On-Demand Reasoning.} This process is event-driven. It remains dormant to conserve resources and activates instantaneously upon receiving a trigger to generate responses, referencing the latest state snapshot without contention.

\noindent$\bullet$ \textit{Sleep-Wake Mechanism.}
To minimize computational overhead during idle periods, we incorporate a Sleep-Wake Mechanism. The cloud reasoning model defaults to a \textit{sleep mode}, remaining suspended and occupying negligible resources. The process transitions to the \textit{active mode} instantaneously upon receiving a trigger from the edge. Upon activation, \method{} initiates the following reasoning pipeline to generate responses based on current memory.

\noindent$\bullet$ \textit{Graph-Augmented Reasoning.}
\method{} executes a four-step workflow to generate the response $R$:

\noindent\textbf{i}) \textit{Query Vectorization}: The user query $Q$ is encoded into a vector $\bm{V}_q$ with the identical embedding model $\mathcal{E}_{txt}$ employed in the graph construction phase:
    \begin{equation}
    \bm{V}_q = \mathcal{E}_{txt}(Q).
    \label{eq:qv}
\end{equation}

\noindent\textbf{ii}) \textit{Caption Retrieval}: We then calculate the cosine similarity between the query vector $\bm{V}_q$ and the caption vectors $\bm{e}_u$ mounted on each node $u$ in the graph node set $\mathcal{V}$. The nodes associated with the top-$k$ most relevant captions are identified as a semantic anchors set $\mathcal{A}$:
    \begin{equation}
    \mathcal{A} = \texttt{Top-k}(\{u \in \mathcal{V} \mid \cos(\bm{V}_q, \bm{e}_u)\}).
\label{eq:cr}
\end{equation}

\noindent\textbf{iii}) \textit{Subgraph Extraction}: Centered on these anchors, we traverse the graph to extract their first-order neighbors and interconnecting edges. This induces a structure that forms a relevant subgraph $\mathcal{G}_{sub} \subset \mathcal{G}_t$, which retains only the entities and relations directly relevant to the user's query:
\begin{equation}
\begin{aligned}
\mathcal{V}_{sub} &= \mathcal{A} \cup \bigcup_{a \in \mathcal{A}} \mathcal{N}(a), \\
\quad \mathcal{G}_{sub} &= (\mathcal{V}_{sub}, \{(u, v) \in \mathcal{E} \mid u, v \in \mathcal{V}_{sub}\}).
\end{aligned}
\label{eq:se}
\end{equation}

\noindent\textbf{iv}) \textit{Graph Serialization}: The retrieved $\mathcal{G}_{sub}$ is serialized into the JSON format. Finally, the serialized subgraph, the global visual context $\bm{H}_t$, and the original query are jointly fed into the LMM to generate the final response:
    \begin{equation}
    R = \texttt{LMM}(\bm{H}_t, \texttt{Serialize}(\mathcal{G}_{sub}), Q).
    \label{eq:gs}
\end{equation}

\noindent\textbf{Merits.} The multi-process asynchronous architecture decouples the memory management from reasoning, ensuring that continuous graph updates do not block instantaneous user interactions. Graph-augmented retrieval further reduces overhead by filtering out irrelevant context, enabling LMM reasoning on a concise subgraph instead of the full history. Collectively, these optimizations ensure responsiveness, reducing end-to-end latency to 2.99s, significantly outperforming cloud-only baselines (4.08s) while remaining comparable to local-only baseline (2.72s, \textit{cf.} 
 Table~\ref{tab:main_results}).

\subsection{Implementation Details}

\noindent\textbf{Edge Node.}
Visual Feature Extraction (Eq.~\ref{eq:vis_enc}) adopts the vision encoder (0.4B) from \textit{VideoLLaMA3}-\textit{7B}~\citep{videollama3}, and Semantic Caption Generation (Eq.~\ref{eq:captions}) is powered by \textit{Qwen3}-\textit{VL}-\textit{2B}-\textit{Instruct}~\citep{qwen3}. 

\noindent\textbf{Cloud Server.}
The Graph Builder employs \textit{SpaCy} for syntactic parsing, \textit{NLI}-\textit{Deberta}-\textit{v3}~\citep{deberta} for entailment verification (Eq.~\ref{eq:ev}), and \textit{Qwen3}-\textit{Embedding}-\textit{0.6B}~\citep{qwen3} for caption vectorization (Eq.~\ref{eq:cv}). The Inference Engine is based on \textit{VideoLLaMA3}-\textit{7B}~\citep{videollama3}. The following hyperparameters are adopted: the keyframe selection buffer is $N_w=64$, and the edge node condenses every $N_{temp}=16$ raw frame features into $N_{target}=2$. The visual compression bank size $N_{bank}$ is 128. For subgraph extraction, \method{} retrieves the top-$3$ most relevant captions.

\begin{table*}[t] \small
\centering

\setlength{\tabcolsep}{12pt}
\begin{tabular}{lcccccc}
\toprule
\multirow{2}{*}{\textbf{Baselines}} & \multicolumn{1}{c}{\textbf{LVBench}} & \multicolumn{1}{c}{\textbf{VideoMME}} & \multicolumn{1}{c}{\textbf{RTV-Bench}} & \multicolumn{3}{c}{\textbf{Efficiency Metrics}} \\
\cmidrule(lr){2-2} \cmidrule(lr){3-3} \cmidrule(lr){4-4} \cmidrule(lr){5-7}
 & \textbf{Acc. (\%) $\uparrow$} & \textbf{Acc. (\%) $\uparrow$} & \textbf{Acc. (\%) $\uparrow$} & \textbf{Comm. (MB) $\downarrow$} & \textbf{Comp. (\%)} $\uparrow$ & \textbf{Lat. (s) $\downarrow$} \\
\midrule
\texttt{LO} & 32.42\small${\textcolor{gray}{\pm0.05}}$ & 34.11\small${\textcolor{gray}{\pm2.33}}$ & 30.44\small${\textcolor{gray}{\pm0.12}}$ & \textbf{0.00} & -- & \textbf{2.72} \\
\texttt{CO} & \textbf{39.23\small${\textcolor{gray}{\pm0.50}}$} & \textbf{46.67\small${\textcolor{gray}{\pm0.26}}$} & \textbf{35.39\small${\textcolor{gray}{\pm0.27}}$} & 2549.78 & -- & 4.08 \\
\texttt{EC} & 36.59\small${\textcolor{gray}{\pm0.12}}$ & 43.33\small${\textcolor{gray}{\pm0.67}}$ & 34.55\small${\textcolor{gray}{\pm0.40}}$ & 637.45 & \underline{75.00} & 4.09 \\
\rowcolor[rgb]{ .930,  .980,  .980} 
\midrule
\textbf{\method{}} & \underline{38.93\small${\textcolor{gray}{\pm0.38}}$} & \underline{44.11\small${\textcolor{gray}{\pm0.23}}$} & \underline{34.96\small${\textcolor{gray}{\pm0.31}}$} & \underline{316.42} & \textbf{87.59} & \underline{2.99} \\
\bottomrule
\end{tabular}
\caption{Main results on LVBench, VideoMME-Long, and RTV-Bench (\S\ref{exp:main}). Efficiency Metrics uses LVBench as the representative example. ``Comp.'' denotes the data compression rate relative to the \texttt{CO}. ``Lat.'' measures the end-to-end latency time, spanning from the user's query to the response. Bold denotes the best result, \underline{underlined} denotes the second-best.}
\label{tab:main_results}

\end{table*}

\section{Experiment}
\label{sec:exp}
\subsection{Experimental Setting}
\label{sec:setting}
\noindent{\noindent\textbf{Datasets.} To evaluate \method{} in realistic streaming scenarios, we employ three benchmarks:}

\noindent$\bullet$ \textit{VideoMME}-\textit{Long}~\citep{videomme}: A comprehensive benchmark spanning 6 primary visual domains with 30 subfields designed to evaluate multi-modal reasoning and cross-domain generalization. The subset, with an average duration of 3,160 seconds, contains 900 single-choice questions that require temporal reasoning and long-term context understanding across varying video lengths.

\noindent$\bullet$ \textit{LVBench}~\citep{lvbench}: An extremely long video understanding benchmark, designed to assess long-term memory and ``needle-in-a-haystack'' retrieval through single-choice QA pairs. A specific subset of videos with durations exceeding 2,800 seconds is adopted to measure the ultra-long video understanding capability (average duration of 5,018 seconds), comprising a total of 1,007 questions.

\noindent{$\bullet$ \textit{RTV-Bench}~\citep{rtv}: A real-time video benchmark designed for dynamic streaming tasks. It includes 167.2 hours of streaming-style videos, averaging 1,092 seconds per session, across domains such as egocentric intelligence, intelligent driving, and sports analytics.}

\noindent\textbf{Baselines.} \!We compare \method{} with three baselines:

\noindent$\bullet$  Local-Only (\texttt{LO}): Powered by the lightweight \textit{VideoLLaMA3}-\textit{2B}~\citep{videollama3}, it processes all video frames locally and establishes the lower bound for task accuracy due to limited model capacity, while representing the upper bound for communication efficiency.

\noindent$\bullet$  Cloud-Only (\texttt{CO}): \textit{VideoLLaMA3}-\textit{7B}~\citep{videollama3} is adopted to process raw video frames uploaded from the edge at 1 FPS and serves as the accuracy upper bound, but incurs the maximum communication overhead.

\noindent$\bullet$ Edge-Cloud Hybrid (\texttt{EC}): A collaborative baseline designed to reduce bandwidth consumption. The edge samples frames at 1 FPS and performs 4$\times$ downsampling before transmission. The cloud applies a latent diffusion model (\textit{AuraSR}-\textit{v2}) to restore resolution for inference.

\subsection{Main Results}
\label{exp:main}
{Table~\ref{tab:main_results} summarizes the performance across the main benchmarks. On VideoMME-Long, \method{} (44.1\%) gains \textbf{+10.0\%} over \texttt{LO} (34.1\%) and retains \textbf{94.5\%} of \texttt{CO}'s performance (46.7\%). On LVBench, \method{} (38.9\%) outperforms \texttt{LO} (32.4\%) by \textbf{+6.5\%} and retaining \textbf{99.2\%} of \texttt{CO}'s performance (39.2\%). Regarding latency, \method{} (2.99s) reduces end-to-end latency by \textbf{1.09s} compared with \texttt{CO} (4.08s), while remaining close to \texttt{LO} (2.72s). This efficiency stems from our asynchronous design (\S\ref{method:cloud}), where the cloud memory is constructed in real time and requires only \textbf{0.16s} to retrieve the subgraph for inference.}
{On RTV-Bench~\citep{rtv}, \method{} further achieves 35.0\%, compared with \texttt{LO} (30.4\%), \texttt{EC} (34.6\%), and \texttt{CO} (35.4\%), indicating that the same edge-cloud decomposition transfers to streaming perception tasks. We additionally evaluate open-ended QA on VideoSIAH-Eval~\citep{longvt} (Appendix) and test H.264-matched transmission (Appendix.}
For additional details on the latency calculation, please refer to Appendix.

\subsection{Diagnostic Experiment}
\label{exp:ana}

\begin{figure*}[t]
    \centering
\includegraphics[width=0.99\linewidth]{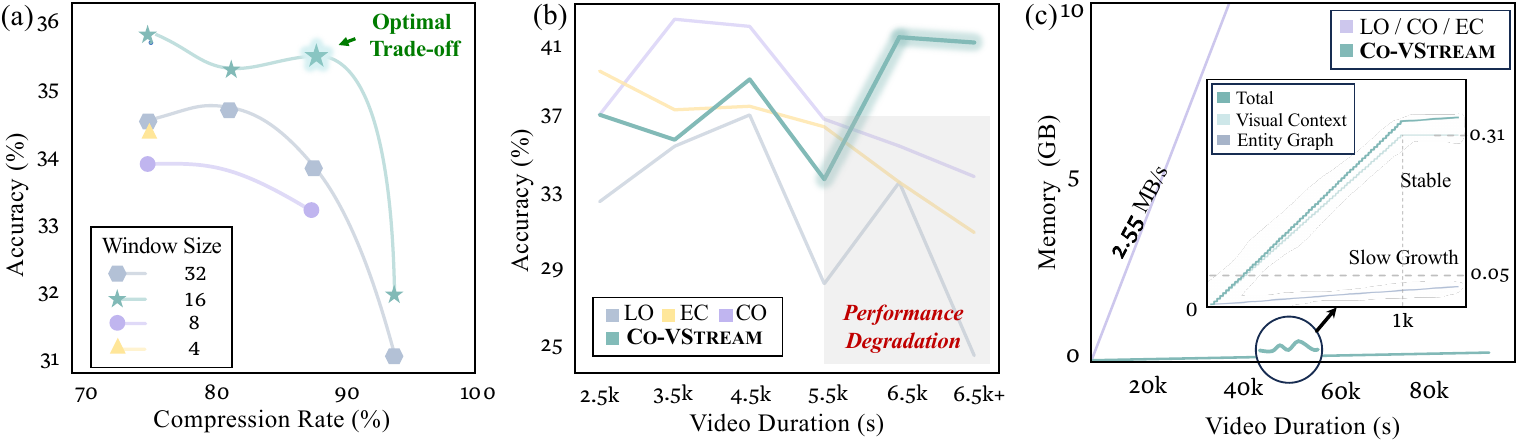}
    \caption{Diagnostic experiments of \method{} performance (\S\ref{exp:ana}).}
    \label{fig:ana}
\end{figure*}

\begin{table}[t] \small 
\centering

\setlength{\tabcolsep}{8pt} 

\definecolor{lightcyan}{rgb}{.930, .980, .980}

\begin{tabular}{ccccc} 
\toprule
\textbf{ID} & \textbf{Components} & \textbf{Acc. ($\uparrow$)} & \textbf{Comm. ($\downarrow$)} & \textbf{Comp. ($\uparrow$)} \\
\midrule
1 & T & 35.45 & 316.41 & 87.59\% \\
2 & S & 36.94 & \textbf{137.62} & \textbf{94.61}\% \\
3 & G & 37.24 & 2590.67 & 0.00\% \\
\midrule
4 & T+G & 38.16 & 316.42 & 87.59\% \\
5 & S+G & 38.23 & 137.63 & 94.60\% \\
6 & T+S & 34.66 & 333.22 & 86.93\% \\
\midrule
\rowcolor{lightcyan} 
7 & T+S+G & \textbf{38.93} & 316.42 & 87.59\% \\
\bottomrule
\end{tabular}
\caption{Ablation study of components. \textsc{T} for Temporal Feature Condensation; \textsc{S} for Adaptive Keyframe Selection; \textsc{G} for Entity Graph Construction (Eq.~\ref{eq:ev}-\ref{eq:cv}).}
\label{tab:ablation}
\end{table}

To provide a deeper understanding of \method{}'s internal mechanisms, we conduct diagnostic analyses:

\noindent\textbf{Key Component Analysis.}
To verify the contribution of each module, we progressively dismantle \method{}. The results are shown in Table~\ref{tab:ablation}. Comparing ID 6 (\textsc{T} + \textsc{S}) and ID 7 (\method{}), the inclusion of the \textsc{G} increases accuracy from 34.66\% to 38.93\%. This suggests that the structured graph compensates for the information loss caused by the compression. 
{Furthermore, it is worth noting that while ID 4 (\textsc{T} + \textsc{G}) and ID 5 (\textsc{S} + \textsc{G}) yield substantial compression rates, their accuracy degrades when failing to construct the graph over keyframes or omitting temporal condensation. Ultimately, by jointly integrating these components (ID 7), \method{} achieves best performance and reduces bandwidth by approximately 88\% compared to \texttt{CO}.}

\noindent\textbf{Compression Strategy Trade-off}.
{To evaluate how window size and output frame count affect Temporal Feature Condensation, we test on the \textit{full} LVBench and find two key trends, as shown in Fig.~\ref{fig:ana}a. Regarding the window size, a setting of $\mathtt{16}$ frames yields higher performance than both larger ($\mathtt{32}$) and smaller ($\mathtt{8}$) windows. This indicates that an overly large window tends to blend distinct semantic events, whereas a smaller window fragments context. Regarding the compression rate, within the optimal window size ($\mathtt{16}$), reducing the compression rate from 87.5\% ($\mathtt{16} \to \mathtt{2}$) to 75.0\% ($\mathtt{16} \to \mathtt{4}$) yields only a marginal accuracy gain (+0.28\%) but doubles the bandwidth cost. Meanwhile, extreme compression (93.8\%, $\mathtt{32} \to \mathtt{2}$) causes a significant performance drop to 31.16\%. Consequently, the 87.5\% setting ($\mathtt{16} \to \mathtt{2}$) is adopted as the standard configuration. A finer bank-capacity ablation is deferred to Appendix.}

\noindent\textbf{Temporal Robustness Analysis}.
We categorize the LVBen-ch into 1,000-second intervals (starting from 1,000s) to evaluate \method{}'s robustness over extended durations.
As illustrated in Fig.~\ref{fig:ana}b, all baselines exhibit a downward trend for durations \textgreater 5,000s. Conversely, \method{} demonstrates an upward trend starting from the 4,000s' mark. Notably, in the ultra-long category (\textgreater 6,000s), \method{} achieves 42.92\% accuracy, surpassing \texttt{CO} (35.00\%) by \textbf{7.92\%}.
This divergence highlights the robustness of \method{} when processing ultra-long videos. LMMs are constrained by context windows, which restrict their handling of long videos and lead to severe information loss and escalating computational costs. In contrast, \method{} maintains a low-cost structured graph. Through query-driven retrieval, it identifies the most relevant subgraph, preserving critical information without overloading the inference engine.

\noindent\textbf{Memory Scalability Analysis}.
{While existing datasets are limited to hours-long videos, which are far shorter than true always-on scenarios, we assess \method{}'s feasibility for 24/7 deployment by proactively simulating a continuous video stream of 90,000 seconds (exceeding 24 hours) and monitoring the memory footprint on the cloud server.}
As shown in Fig.~\ref{fig:ana}c, baselines require storing all historical frames. resulting in a steep linear memory expansion (2.55 MB/s), accumulating to 229.5 GB after 90,000s. Such unbounded growth makes long-term deployment computationally prohibitive. In contrast, \method{} exhibits a saturation behavior, maintaining a low memory footprint totaling only 438.3 MB, a \textbf{523$\times$} reduction compared to baselines, at the end of the simulation. Fig.~\ref{fig:ana}c provides a zoomed-in view of our memory dynamics: the Global Visual Context grows initially, but will hit the pre-defined capacity limit, stabilizing at approximately 316 MB. Once saturated, the only growing component is the Entity Graph, which consumes merely \textbf{122.3 MB} to represent the entire 24-hour stream. {We further profile payload formatting in Appendix.}

\noindent{\textbf{Hardware Deployment Analysis}.
All our experiments are conducted on an NVIDIA A40 (48GB). We further stress-test the edge pipeline on the RTX 3090 (24GB) and RTX 2080 (8GB). Table~\ref{tab:hardware} shows that accuracy remains stable across the three hardware profiles. Since \method{} decouples perception from reasoning, the edge runs only the lightweight 0.4B vision encoder and 2B captioner, which fit within 8GB without performance loss. The 5.9 TFLOPs requirement is measured in Dense BF16 precision. For broader context, NVIDIA Jetson Thor delivers 162.3 TFLOPS of dense FP16 tensor-core performance, and we also provide a real-phone demo video in the supplementary material.}
\begin{table}[t]  \small
\centering

\setlength{\tabcolsep}{8pt} 

\definecolor{lightcyan}{rgb}{.930, .980, .980}

\begin{tabular}{lccc} 
\toprule
\textbf{Device} & \textbf{Memory} & \textbf{FP32 TFLOPS} & \textbf{Acc. (\%)} \\
\midrule
A40 & 48 GB & 37.42 & 38.93 \\
RTX 3090 & 24 GB & 35.58 & 38.63 \\
\cellcolor{lightcyan}\textbf{RTX 2080} & \cellcolor{lightcyan}\textbf{8 GB} & \cellcolor{lightcyan}\textbf{10.10} & \cellcolor{lightcyan}\textbf{39.32} \\
\bottomrule
\end{tabular}
\caption{Edge-device stress test. Accuracy remains stable across three hardware profiles.}
\label{tab:hardware}
\end{table}

\subsection{Qualitative Analysis}
\label{exp:case}

\begin{figure*}[t]
    \centering
\includegraphics[width=0.9\linewidth]{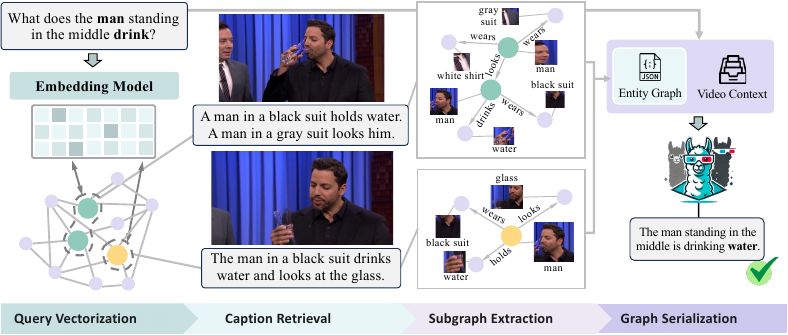}
    \put(-375,6.8){\scriptsize
    \textbf{
    (Eq.~\ref{eq:qv})}}
    \put(-261.8,7.2){\scriptsize
    \textbf{
    (Eq.~\ref{eq:cr})}}
    \put(-135.2,7.2){\scriptsize
    \textbf{
    (Eq.~\ref{eq:se})}}
    \put(-27.7,7.2){\scriptsize
    \textbf{
    (Eq.~\ref{eq:gs})}}
    \caption{Demonstration of Graph-Augmented Reasoning
Pipeline's workflow (\S\ref{exp:case}).}
    \label{fig:case}
\vspace{-13pt}
\end{figure*}
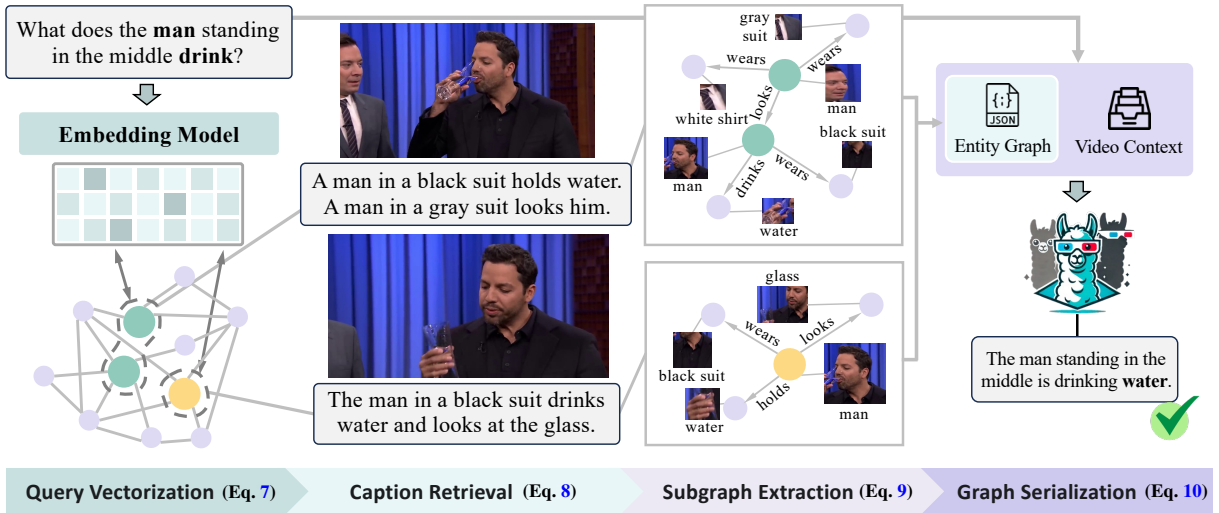

To intuitively demonstrate the Graph-Augmented Reasoning Pipeline, we visualize the workflow as a step-by-step process in Fig.~\ref{fig:case}.
The user's question (\eg, ``What does the man standing in the middle drink?'') is first encoded into a vector. Based on cosine similarity, \method{} retrieves the Top-3 captions that are most similar to the query (\eg, ``A man in a black suit holds water...'').
Since these captions are mounted directly on the corresponding entities, \method{} identifies these entity nodes (\eg, Man, Black Suit) and their 1-hop neighbors, thus extracting a minimal relevant subgraph.
Finally, the subgraph is serialized into a structured JSON format and injected into the original input prompt as graph context to generate the response.

\section{Related Work}

\noindent\textbf{Long-term Video Understanding}.
Enhancing LMMs' capabilities for long video understanding attracts widespread community attention. These approaches can be broadly categorized into three main streams.
The first category focuses on context scaling and architectural adaptation. Some work on scaling training infrastructure~\citep{LWM, LongVILA}. Others focus on efficient long-context modeling~\citep{LLaMA-VID, Long-Context-Transfer, LongVLM, Video-CCAM}.
The second category is memory-augmented streaming and compression. These approaches manage memory banks~\citep{MovieChat, MA-LMM, StreamForest, Flash-VStream}, compress features~\citep{LongVU, LIGHTMEM, Video-XL, SLV,  MovieChat+}, or select visual subsets~\citep{SEAL}.
The third category is agentic reasoning and structured retrieval. These methods leverage LMMs as planners~\citep{VideoAgent, VideoAgent2, AVUA, ViTCoT, HMP}. Retrieval-augmented generation based frameworks index video content~\citep{Goldfish, VideoRAG,VideoAgent} or convert videos into textual logs or semantic graphs~\citep{VideoMindPalace, GraphVideoAgent, LifelongMemory}.

\noindent Despite impressive, existing methods require \textit{offline processing of entire videos}. This reliance causes latency bottlenecks and high bandwidth costs, rendering them impractical for constrained devices that require \textit{immediate} responsiveness. \method{} instead treats video as \textit{continuous streams} and operates as an edge-cloud collaborative framework. The edge condenses the video streams to minimize bandwidth, while the cloud concurrently manages memory and executes reasoning, guaranteeing low-latency responsiveness.

\noindent\textbf{Collaborative Cloud-Edge Intelligence.}
Deploying LMMs in resource-constrained edge nodes also receives widespread attention from the community. These approaches can be broadly categorized into three main streams.
The first category focuses on the dynamic inference offloading~\citep{1, 2, 7, 8, 14,  17, 18,19}. These methods decompose the inference process to balance latency and computational consumption, \eg, by routing tokens~\citep{1, 2, 9, 19}, partitioning model layers~\citep{7}, or employing different decision-making algorithms~\citep{8,14, 17,18}.
The second category is cloud-driven model adaptation~\citep{3, 4, 6, 10, 12, 19}. These methods leverage the cloud models to enhance the edge performance, \eg, distilling knowledge~\citep{12, 19} or generating codes for the edge nodes~\citep{4, 10}.
The third category is context-aware collaborative generation~\citep{5, 11, 13, 15, 20}. In text domains, methods either retrieve cloud-hosted summaries or perform local prompt completion. In visual domains, common methods include edge-based filtering~\citep{11,20} or using cloud inference as historical context to guide edge predictions.

\noindent{While edge-cloud video analysis is well-established, existing methods~\citep{re0,re1,re2} predominantly tackle low-level vision tasks (\eg, frame tagging, motion detection, and object counting) rather than high-level video understanding. Unlike these methods, \method{} enables video understanding (\eg, localization, summarization, and open-ended QA). Specifically, by decoupling edge-side semantic condensation from cloud-side memory maintenance, \method{} ensures low-latency execution and low bandwidth usage for streaming videos.}

\section{Conclusion}

{This work bridges the disparity between the limited hardware capacity of edge devices and the computational demands of long video understanding. We propose \method{}, a collaborative framework that reconciles bandwidth and reasoning constraints by decoupling edge-side condensation cloud-side memory and reasoning. We hope this work directs community attention toward the domain of always-on deployment, thereby extending the research scope of video understanding beyond offline benchmarks to real-world streams.}

\newpage
\clearpage
{\small
\bibliography{aaai2027}
}

\end{document}